\pdfoutput=1 
\documentclass[10pt,twocolumn,letterpaper]{article}

\usepackage[pagenumbers]{cvpr} 

\usepackage[dvipsnames]{xcolor}

\usepackage{multirow}
\usepackage{tabularx}
\usepackage{subcaption}
\usepackage{pifont}
\usepackage{array}
\usepackage{makecell}  
\usepackage{multicol}
\usepackage{multirow}
\definecolor{cvprblue}{rgb}{0.21,0.49,0.74}
\usepackage[pagebackref,breaklinks,colorlinks,citecolor=cvprblue]{hyperref}
\usepackage{graphicx}
\usepackage{amsmath}
\usepackage{amssymb}
\usepackage{booktabs}
\usepackage[normalem]{ulem}
\usepackage{float}
\useunder{\uline}{\ul}{}
\def\paperID{10017} 
\def\confName{CVPR}
\def\confYear{2025}

\newcommand\blfootnote[1]{%
  \begingroup
  \renewcommand\thefootnote{}\footnote{#1}%
  \addtocounter{footnote}{-1}%
  \endgroup
}

\title{CompetitorFormer: Competitor Transformer for 3D Instance Segmentation}

\author{Duanchu Wang$^{1}$ \quad Jing Liu$^{2}$\quad Haoran Gong$^{2}$\quad Yinghui Quan$^{1}$\quad Di Wang$^{2\dagger}$\\$^{1}$School of Electronic Engineering, Xidian University \\ $^{2}$School of Software Engineering, Xi’an Jiaotong University \\ E-mails: \texttt{\{wangduanchu\}@stu.xidian.edu.cn} }

\begin{document}
\maketitle

\blfootnote{*Corresponding authors: Di Wang}

\begin{abstract}
Transformer-based methods have become the dominant approach for 3D instance segmentation. 
These methods predict instance masks via instance queries, ranking them by classification confidence and IoU scores to select the top prediction as the final outcome.
However, it has been observed that the current models employ a fixed and higher number of queries than the instances present within a scene. In such instances, multiple queries predict the same instance, yet only a single query is ultimately optimized.
The close scores of queries in the lower-level decoders make it challenging for the dominant query to distinguish itself rapidly, which ultimately impairs the model's accuracy and convergence efficiency. This phenomenon is referred to as inter-query competition.
To address this challenge, we put forth a series of plug-and-play competition-oriented designs, collectively designated as the \textbf{CompetitorFormer}, with the aim of reducing competition and facilitating a dominant query.
Experiments showed that integrating our designs with state-of-the-art frameworks consistently resulted in significant performance improvements in 3D instance segmentation across a range of datasets.
\end{abstract}    

\section{Introduction}
\label{sec:intro}

The landscape of modern 3D instance segmentation methods has changed dramatically following the pioneering work of 2D segmentation transformer, Mask2Former \cite{mask2former}. The transformer-based methods \cite{spformer,mask3d,maft,queryformer,al20233d,oneformer3d} have demonstrated superior performance compared to the top-down proposal-based methods \cite{3dbot,3dsis,3dmpa,td3d,sun2023neuralbf,GSPN,sphericalmask} and bottom-up grouping-based methods \cite{pointgroup,hais,He2021dyco3d,SSTNet,Maskgroup,softgroup}. This is attributed to their comprehensive end-to-end pipeline architecture that directly outputs instance predictions. Fundamentally, these models leverage a predefined number of instance queries and refine mask predictions via capturing global feature.

\begin{figure}[t]
	\centering
	\begin{subfigure}{0.49\linewidth}
		\centering
		\includegraphics[width=0.49\columnwidth]{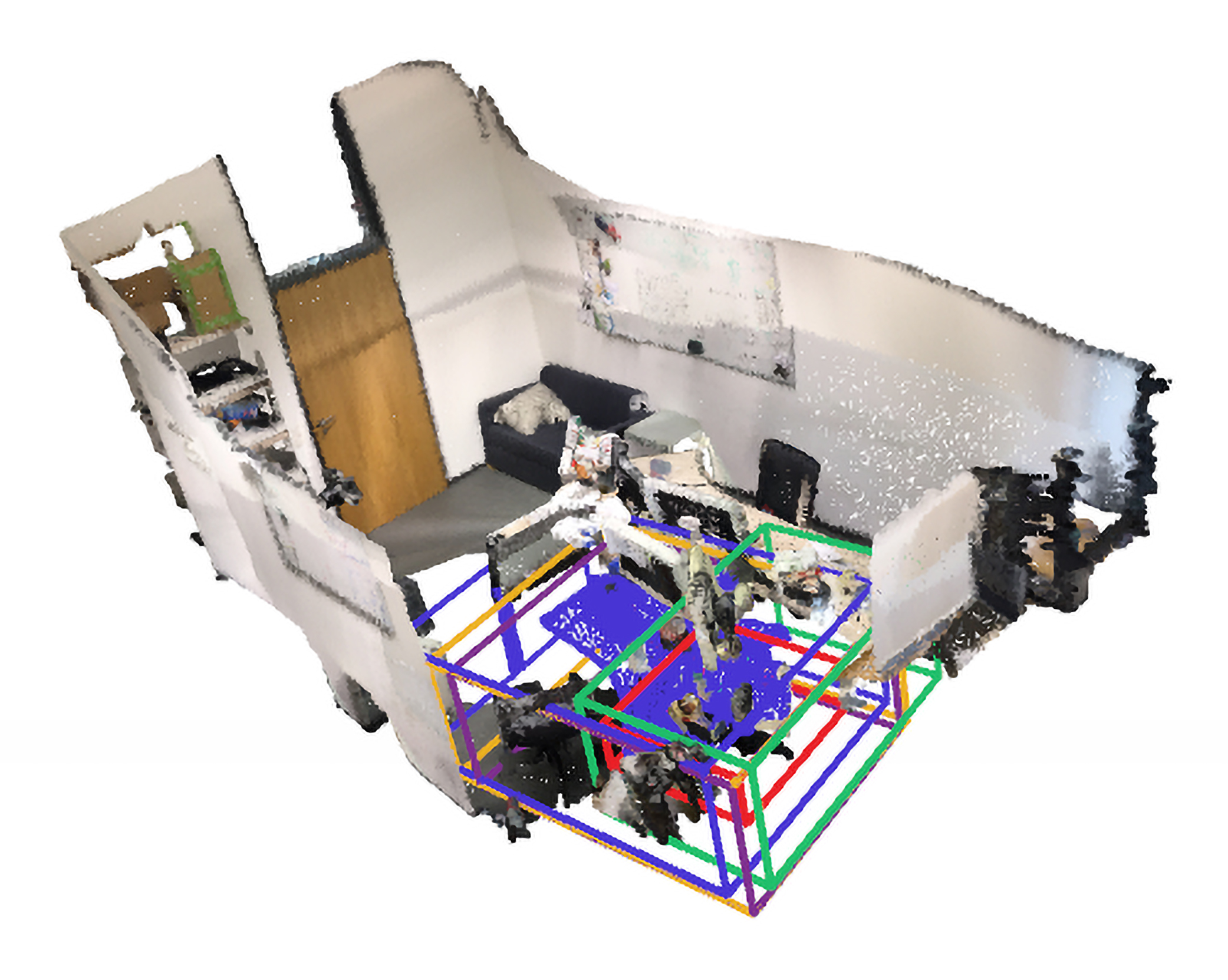}
		\includegraphics[width=0.49\columnwidth]{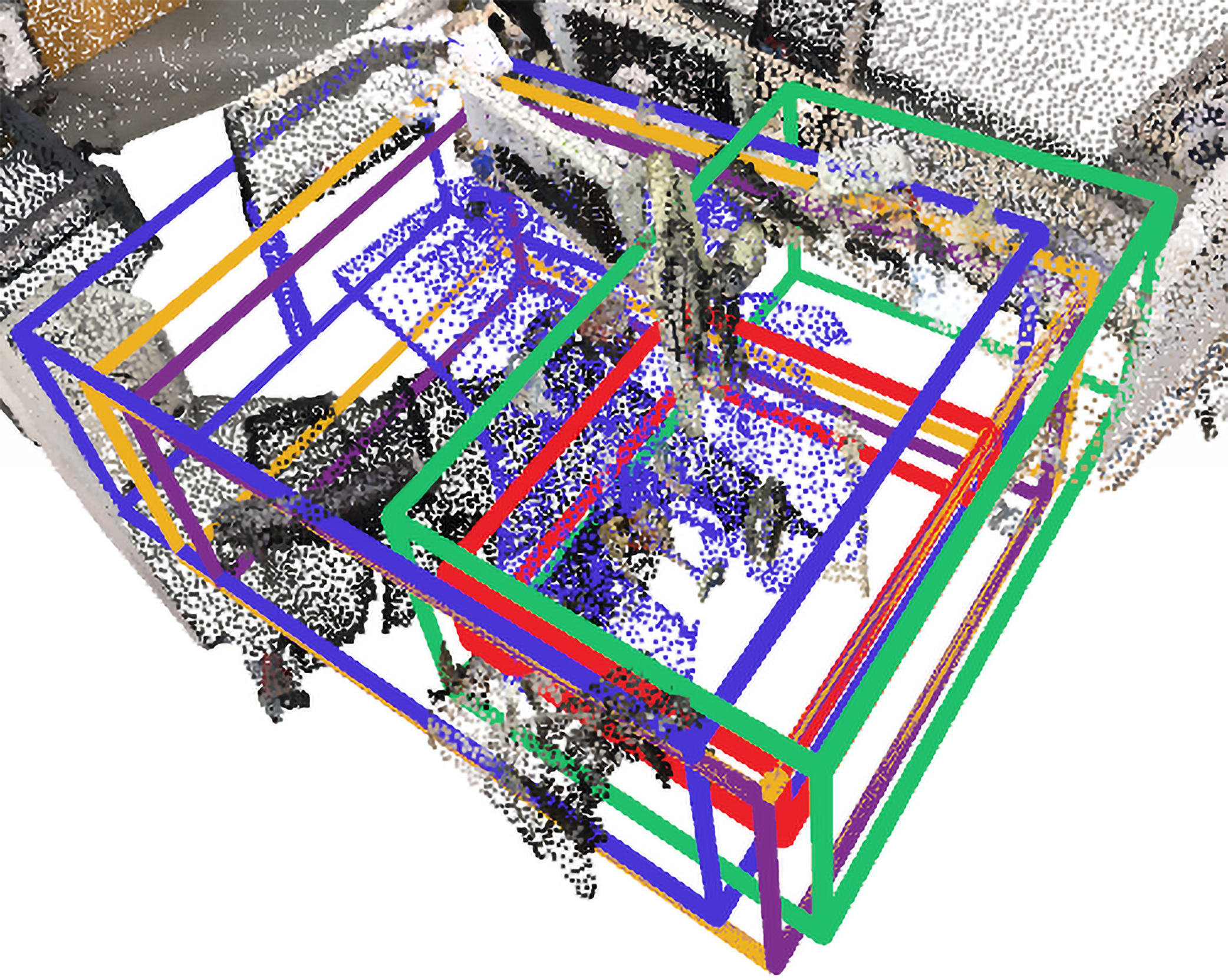} 
		\includegraphics[width=0.49\columnwidth]{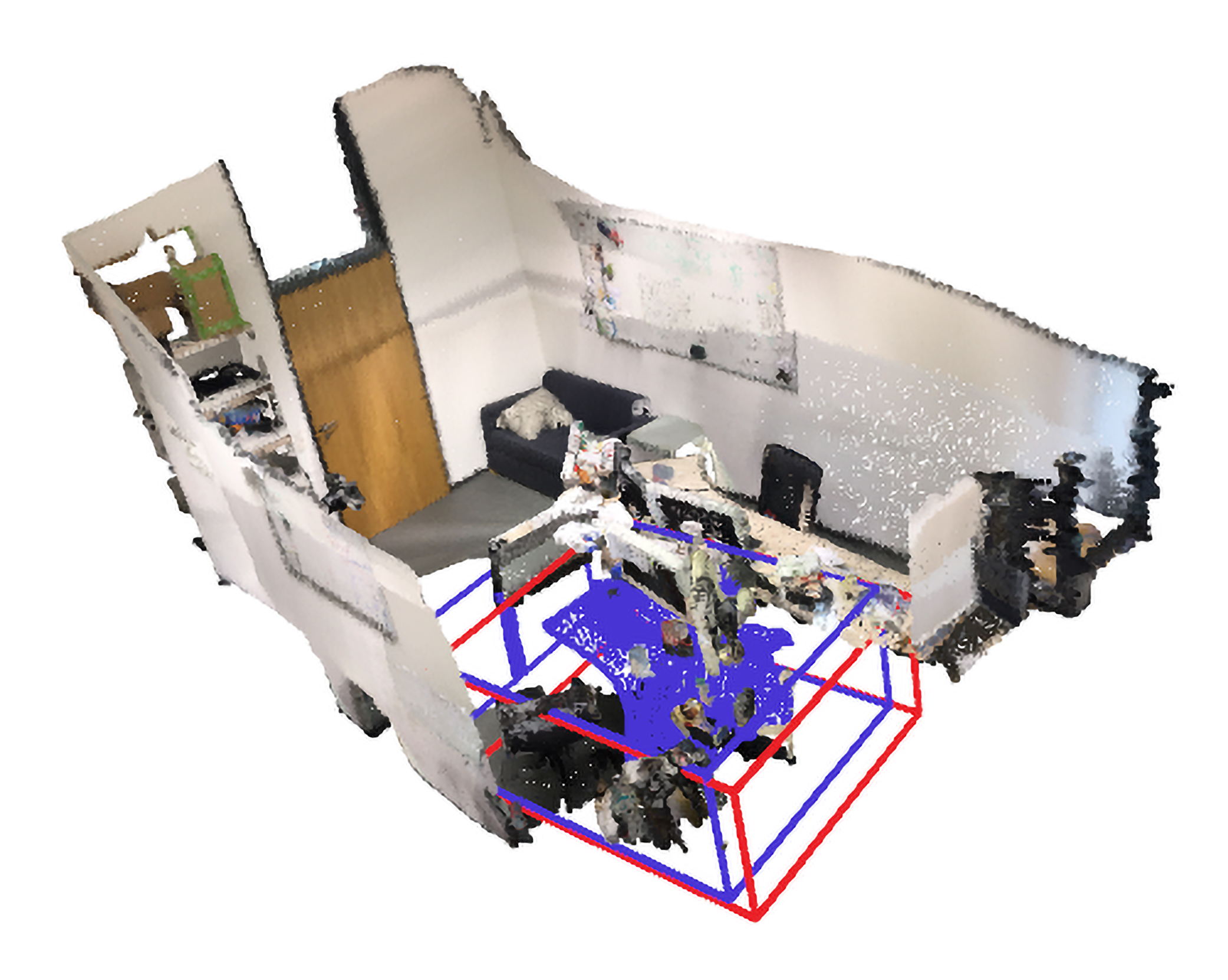} 
		\includegraphics[width=0.49\columnwidth]{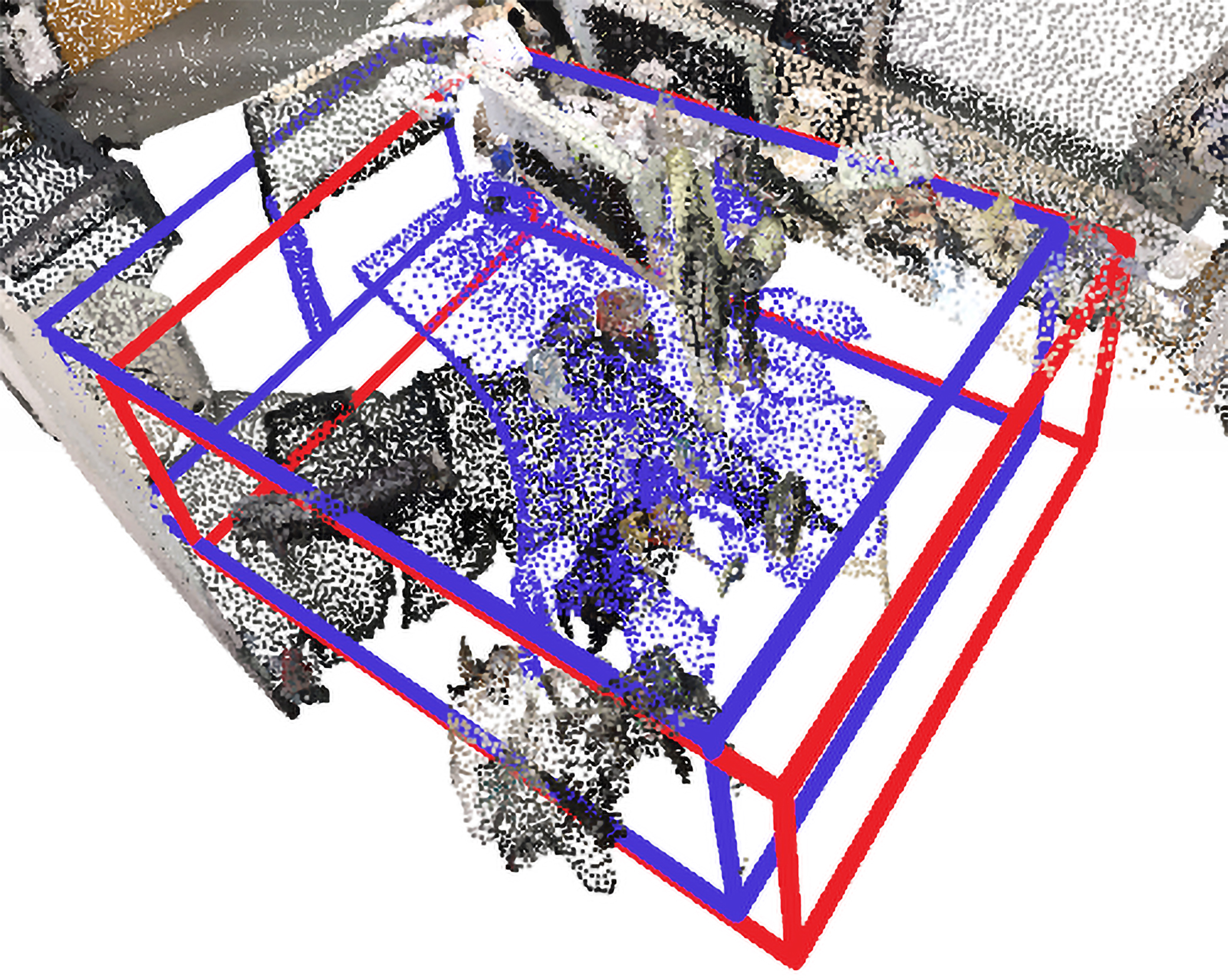} 
		
		\caption{}
		\label{fig:1-1}
	\end{subfigure}
	\begin{subfigure}{0.49\linewidth}
		\centering
		\includegraphics[width=\columnwidth]{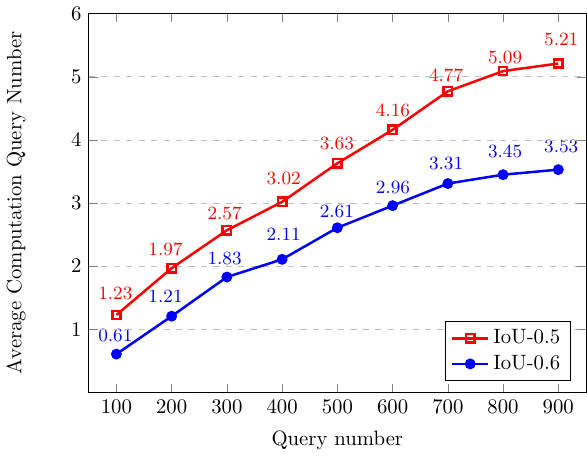}
		\caption{}
		\label{fig:1-2}
	\end{subfigure}
	\vspace{-8pt}
	\caption{(a) The visualization of competing queries from the initial decoder layer of SPFormer \cite{spformer}. The 3D bounding box represents the predicted instance. The blue color represents the ground truth, and the remaining colors (e.g., red, green, etc.) represent the prediction boxes. (b) Statistics of the average number of competing queries for SPFormer in the initial decoder layer under different numbers of queries and different IoU thresholds.}
	\label{figure1}
\end{figure}

The majority of current transformer-based methods utilize a fixed number of instance queries that exceeds the actual number of instances, which make multiple queries predict the identical instance. However, only one query can be distinguished, optimized by bipartite matching, or represented an instance in the inference phase. 
The recently proposed EASE-DETR \cite{EASE-DETR} for 2D detection has demonstrated that there is competition between queries that affects the model's detection performance. A similar issue is observed in the domain of 3D instance segmentation. Figure \ref{fig:1-1} illustrates the competing queries from the initial decoder layers of SPFormer \cite{spformer}, while Figure \ref{fig:1-2} provides a quantitative representation of the average number of competing queries. To mitigate such competition, it is ideal to expand the query score by capturing the spatial and competitive relationships between queries, which allows dominant queries to emerge and be distinguished quickly.

The complexity of 3D spatial information is typically greater than that of 2D scenarios. A strategy that relies exclusively on spatial information and competitive relationships (such as EASE-DETR) may prove inadequate for accurately assessing the competitive state between queries. The relational and semantic information between query categories that have been overlooked can be equally beneficial in the construction of competitive states.
Therefore, in this study, we use 3D spatial data and queries, as well as fully incorporate relational and semantic information between categories of queries, in order to further mitigate competition and enhance the segmentation capability of the model.

Specifically, we introduce three novel competition-oriented designs: query competition layer, relative relationship encoding and rank cross attention. These designs effectively utilize the spatial and competitive relationships between queries, and semantic information to mitigate competition. 
First, we determine the spatial and competitive relationships between queries before each decoder layer. The query competition layer utilizes two distinct sets of static embeddings to capture these relationships and fuses them with the instance query semantic feature. This approach enhances classification confidence for matched queries in bipartite matching and attenuates classification confidence for unmatched queries.
Second, the relative relationship encoding refines the weight by quantifying these relationships in conjunction with the query and the key features from self-attention. 
Third, the rank cross attention mechanism amplifies the disparity between queries by normalizing the dot product similarity between each feature and all queries. Overall, our competition-oriented designs improve model's segmentation performance by progressively mitigating inter-query competition.

Our contributions can be summarized as follows:
\begin{itemize}
	\item[$\bullet$] 
	We observe that existing 3D instance segmentation transformer-based methods suffer from inter-query competition, which causes training difficulty, lower accuracy. 
	\item[$\bullet$] 
	To mitigate inter-query competition, we propose three competition-oriented designs, query competition layer, relative relationship encoding, and rank cross attention, collectively referred to as \textbf{CompetitorFormer}.
	\item[$\bullet$] 
	The experimental results demonstrate that CompetitorFormer achieves performance improvements on a variety of popular baselines over multiple datasets, including ScanNetv2, ScanNet200, S3DIS and STPLS3D, ultimately exceeding their results.
\end{itemize}

\section{Related Work}
In this section, we briefly overview related works on 3D instance segmentation, competing mechanism and the clustering of cross-attention.
\subsection{3D instance segmentation.}
Existing works on 3D instance segmentation can be classified into proposal-based, grouping-based and transformer-based. The proposal-based methods \cite{3dbot, 3dsis,3dmpa,td3d,sun2023neuralbf,GSPN} are based on the advancements in 2D image instance segmentation. They start with coarse detection and then perform a refinement step to obtain a detailed segmentation. The essence of these approaches is the fine-grained segmentation of objects within a predicted 3D bounding box. Enhancing the precision of 3D bounding box predictions is a central optimization objective for these approaches.

In contrast, grouping-based methods \cite{pointgroup,hais,He2021dyco3d,SSTNet,Maskgroup,softgroup} adopt a bottom-up pipeline that learns potential embeddings to facilitate point-wise predictions. These methods utilize the centroid offset of each point to predict a shifted point cloud, which is then used in conjunction with the original point cloud to form clusters. Such methods necessitate accurate prediction of the centroid offset for each point, which can be challenging to achieve when considering a diverse range of spatial distributions.

Recently, transformer-based methods \cite{spformer,mask3d,maft,queryformer,al20233d,oneformer3d} have emerged and are rapidly becoming the new state-of-the-art (SOTA). Compared to previous approaches, these methods offer an elegant pipeline that directly outputs instance predictions. The generation of the final instance masks is achieved through the computation of the dot product similarity between queries and superpoint (or voxel) features. Although the powerful architectural advantage has driven the performance of transformer-based approaches, slow convergence and how to mitigate competition queries remain challenges.

\begin{figure*}[ht] 
	\centering 
	\includegraphics[width=0.9\linewidth]{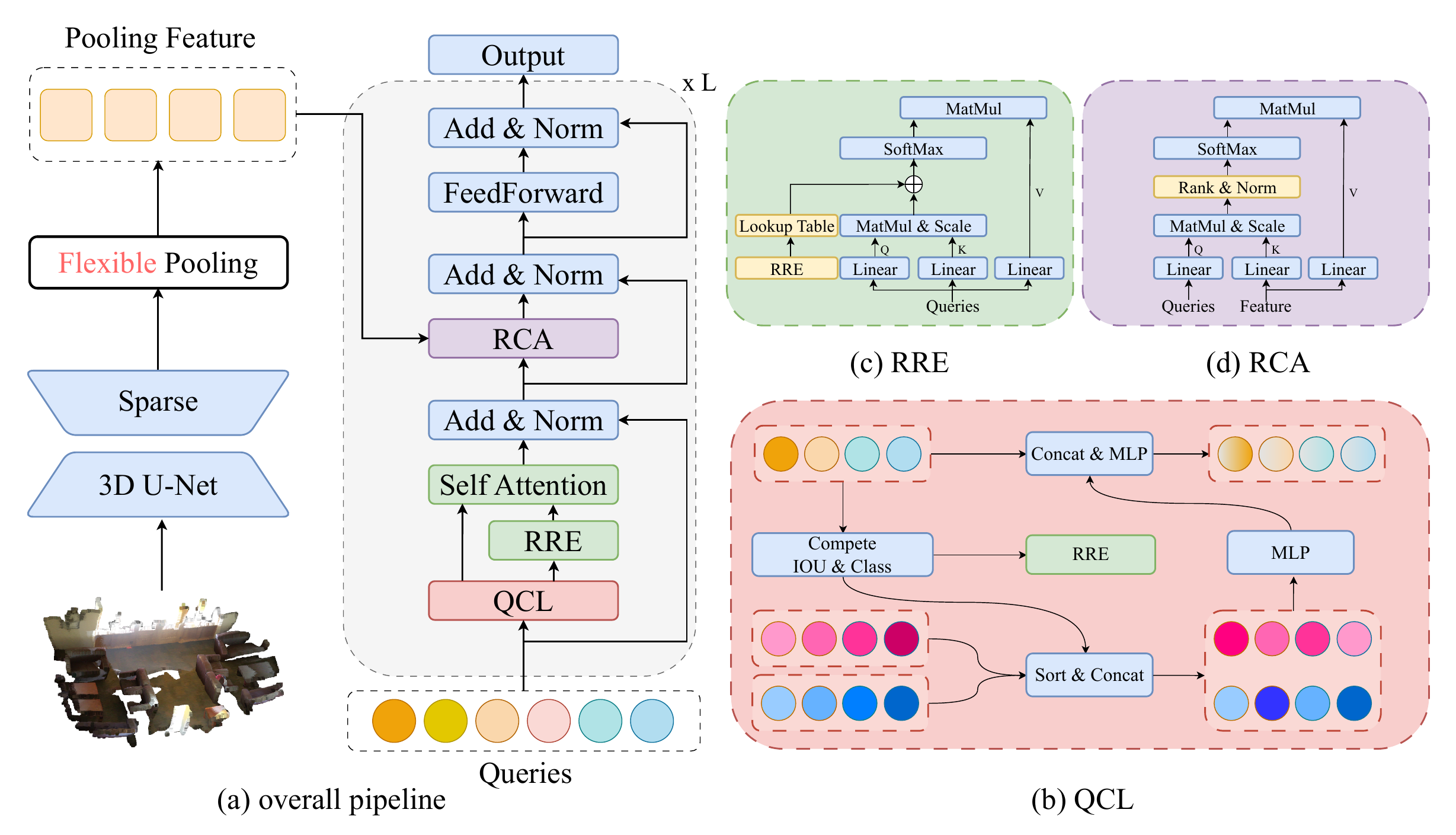} 
	\vspace{-13pt}
	\caption{\textbf{The overview of our pipeline.} (a) shows the overall pipeline, in which the query competitor layer (QCL) processes input to derive spatial and competitive information, bifurcating into branches for static embedding-enhanced instance queries and dynamic self-attention weight adjustment via relative relationship encoding (RRE), followed by the rank cross attention (RCA) for query differentiation. The details of each module are shown in (b), (c), (d), respectively.} 
	\label{figure3} 
\end{figure*}

\subsection{Competing mechanism.} In transformer-based models, the ranking relationship between queries directly affects the final result in the inference phase. Therefore, it is important to formulate a reasonable ranking. The majority of current approaches \cite{Varifocalnet,Tood,GFLoss} rely on the IoU-aware score and joint classification scores, to mitigate ranking inconsistencies and refine detection output accuracy. Additionally, some works \cite{Align-DETR,Rank-DETR} have incorporated IoU-aware techniques to refine the loss and bipartite matching cost designs in DINO-DETR \cite{Dino}. This has resulted in a reduced number of erroneous queries and an acceleration in model convergence. 

Nevertheless, the inter-query competition has been relatively understudied. A recent work of EASE-DETR \cite{EASE-DETR} considered the non-duplicate detection capability to be the result of inter-query competition. It constructed competitive states by sensing the spatial and relative ranking relationships between queries and introduced explicit relations between queries in the decoder to enhance the query suppression process.

\subsection{Clustering cross-attention.} The most popular transformer-based methods use cross-attention to cluster superpoint (or voxel) features into a set of masks, which are progressively refined by several decoder layers. Recently, works like CMT-deeplab \cite{CMT} and KMax-deeplab \cite{KMAX} explored parallels between cross-attention and clustering algorithms in the vision Transformer. Specifically, CMT-deeplab used clustering updates as a complementary term in cross-attention, whereas KMax-deeplab introduced k-means cross-attention, which employs cluster-wise argmax to compute the attention graph and is then directly used for supervision. Another work, QueryFormer \cite{queryformer}, reduced the computational effort by employing the argmax operation along the query dimensions. This approach only requires the computation of an attention graph over a set of analogous backbone features for each query. Furthermore, it is noteworthy that several recent works \cite{xu2022groupvit,liang2023clustseg,mei2024spformer} have similarly revisited the relationship between queries and keys in attention operations. 
The softmax operation was applied along the query dimension in cross-attention, yielding promising results.

\section{Method}
The architecture of the proposed \textbf{CompetitorFormer} is illustrated in Figure \ref{figure3}. We visit the overall pipeline of the modern transformer-based methods in Section \ref{sec:pipeline} and introduce how to capture spatial and competitive relationships between queries in Section \ref{sec:Preparation}. The detailed design of the proposed method, including the competition-oriented  designs, are subsequently illustrated in Section \ref{sec:QCL}, \ref{sec:RRE} and \ref{sec:RCA}.

\subsection{Pipeline} 
\label{sec:pipeline}
\paragraph{Sparse 3D U-Net.} Assuming that an input point cloud contains $N$ points, the input can be formulated as $P \in \mathbb{R}^{N \times 6}$. Each 3D point is parameterized with three colors $r, g, b$, and three coordinates $x, y, z$. Following \cite{Choy_Gwak_Savarese_2019}, we voxelize point cloud, and use a U-Net-like backbone composed of sparse 3D convolutions to extract point-wise features $P' \in \mathbb{R}^{N \times C}$ .
\vspace{-5mm}
\paragraph{Flexible pooling.} Following \cite{oneformer3d}, we implement pooling based on either superpoints or voxels.
Superpoints are oversegmented from input point cloud following \cite{Landrieu_Simonovsky_2018}. Voxels are obtained through voxelization. We refer to this superpoint-based / voxel-based pooling as flexible pooling. The pooling features $F \in \mathbb{R}^{M \times C}$are obtained via average pooling of point-wise features $P' \in \mathbb{R}^{N \times C}$, where $M$ is the number of superpoints or voxels.
\vspace{-5mm}
\paragraph{Query Decoder.}
The initialised query $\mathcal{Q}^0_{N'} \in \mathbb{R}^{N' \times D}$, along with the pooling features $F$, are fed into the decoder layers, where $N'$ and $D$ represent the number of queries and feature dimensions, respectively. The prediction head generates a set of prediction classification scores $\mathcal{P}_{cls}^l \in \mathbb{R}^{N' \times C'}$, prediction IoU scores $\mathcal{S}_{IoU}^l \in \mathbb{R}^{N'}$ and instance masks $\mathcal{M}_{Mask}^l \in \mathbb{R}^{N' \times M}$ based on the updated instance queries, where $l\in\{1,2,\cdots ,L\}$ represents the layer index of the decoder and $C'$ represents the class number.

\subsection{Preparation}  
\label{sec:Preparation}
In the initialisation phase of the model, two sets of static embeddings with the same dimensions as the query are randomly generated for each decoder layer. These are divided into two categories: leader embeddings $E_{Le}^l \in \mathbb{R}^{N' \times D}$ and laggard embeddings $E_{La}^l \in \mathbb{R}^{N' \times D}$. 
The query competition score $\mathcal{K}_{score}^l \in \mathbb{R}^{N'}$ is constructed by multiplying the maximum classification score $\mathcal{P}_{cls}^l$ and the prediction IoU score $\mathcal{S}_{IoU}^l$.
The aforementioned scores $\mathcal{K}_{score}^l$ are employed to calculate the relative difference $\mathcal{C}_{Score}^l \in \mathbb{R}^{N' \times N'}$  between queries, thereby establishing the ``leading/lagging'' relationship $\mathcal{C}_{Rank}^l \in \mathbb{R}^{N' \times N'}$. The formulas are as follows:
\begin{align}
k_i^l &= max(p_i^l) \cdot s_i^l , \\
\mathcal{C}_{Score}^l(i,j) &= k^l_{i}-k^l_{j}, \\
\mathcal{C}_{Rank}^l(i,j) &=  
\begin{cases} 
+1  \quad \mathcal{C}_{Score}^l(i,j)\geq0 \\
-1  \quad \mathcal{C}_{Score}^l(i,j)<0
\end{cases}.
\end{align}

The IoU $\mathcal{C}_{IoU}^l\in \mathbb{R}^{N' \times N'}$ is calculated using the instance mask $\mathcal{M}_{Mask}^l$ predicted by each query, which is then used to facilitate the subsequent steps in identifying competitors. The formula is as follows:
\begin{equation}
\mathcal{C}_{IoU}^l(i,j) = IOU(m_i^l,m_j^l)\text{.}
\end{equation}

\subsection{Query Competition Layer (QCL)} 
\label{sec:QCL}
Here, the query competition layer is introduced before each of the $L-1$ decoder layers. The degree of overlap between queries is indicative of the strength of competition. The indices constituting the set of the most formidable competitors for each query, denoted as $\mathcal{B}^l \in \mathbb{R}^{N'}$, can be obtained from $\mathcal{C}_{IoU}^l(i,j)$. The competitor-query pairs $\mathcal{C}_{c-q}^l$ is jointly constructed by the competitor index $\mathcal{B}^l$ and the ``leading/lagging'' relationship $\mathcal{C}_{Rank}^{l-1}$ as:
\begin{align}
\mathcal{B}^{l-1} &= \overset{n-1}{\underset{j=0}{max}}(\mathcal{C}_{IoU}^{l-1}(i,j))\text{,}\\ 
\mathcal{C}_{c-q}^{l-1} &= \mathcal{C}_{Rank}^{l-1}(i,j)[\mathcal{B}^{l-1}]\text{,} 
\end{align}
where the operation symbol $A[B]$ represents the arrangement of A in accordance with the index of B.

Subsequently, $\mathcal{B}^{l-1}$ and $\mathcal{C}_{c-q}^{l-1}$ are employed in the construction of a list of leader queries $\mathcal{I}_{leader}^l$ and a list of laggard queries $\mathcal{I}_{laggard}^l$ as:

\begin{align}
\mathcal{I}_{leader}^l  & = cat(B_{o}[\mathcal{C}_{c-q}^{l-1}>0],B^{l-1}[\mathcal{C}_{c-q}^{l-1}<0])\text{,} \\
\mathcal{I}_{laggard}^l  & = cat(B_{o}[\mathcal{C}_{c-q}^{l-1}<0],B^{l-1}[\mathcal{C}_{c-q}^{l-1}>0])\text{,} 
\end{align}
where $B_{o}^l$ is a sequence of consecutive integers starting from 1 and ending at $N'$, like $\{1,2,\cdots,N'\}$.

Next, the competition-aware embeddings are constructed based on:
\begin{align}
\hat{E}_{Le}^l = E_{Le}^l[\mathcal{I}_{leader}^l],\quad \hat{E}_{La}^l = E_{La}^l[\mathcal{I}_{laggard}^l],\\
E_{fuse}^l  = MLP(\hat{E}_{La}^l||\hat{E}_{Le}^l ), \\
\hat{\mathcal{Q}}^l  = MLP(\mathcal{Q}^{l-1}||E_{fuse}^l ).
\end{align}

The two static embeddings are then ordered according to the leader and laggard lists and the updated two static embeddings are concatenated in the feature dimension. 
Subsequently, the embeddings are fused back to the original dimensions using a fully-connected layer, resulting in features $E_{fuse}^l$ that encode the competitive relationships between queries. Finally, the aforementioned process is repeated for $E_{fuse}^l$ and $\mathcal{Q}^{l-1}$, thereby completing the update to the query. The key motivation behind QCL is adjusting the classification scores of the instance queries according to their spatial and competitive relationships, integrating semantic features, thereby promoting positive predictions and suppressing negative predictions.

\subsection{Relative Relationship Encoding (RRE)}  
\label{sec:RRE}
In this work, we draw inspiration from MAFT \cite{maft} and adopt a contextual relative relationship encoding approach in self-attention. Our design shares certain similarities with the relative position encoding technique employed in MAFT. However, there is a crucial differentiation exists: MAFT computes relative positions concerning queries and superpoints, whereas our focus lies in encoding the relative relationships between queries themselves. In the following paragraphs, we provide a detailed explanation of the methodology utilized in establishing these inter-query relative relationships.

First, the ``leading/lagging'' relationship $\mathcal{C}_{Rank}^l$ and the degree of competition $\mathcal{C}_{IoU}^l$ are calculated in the \textbf{preparation} phase. 
Subsequently, we further integrate $\mathcal{C}_{Rank}^l$ and $\mathcal{C}_{IoU}^l$. The $\mathcal{C}_{Rank}^l$, being binary, is denoted by $+1$ and $-1$. We directly multiply the $\mathcal{C}_{Rank}^l$ values with $\mathcal{C}_{IoU}^l$, to construct the relative competitive state $R_{state}$ with ``leading/lagging''.
After that, relative competitive state  $R_{state}$ is quantized into discrete integers $\hat{R}_{state}^l$ as:
\begin{align}
R_{state}^l = \mathcal{C}_{Rank}^l \cdot \mathcal{C}_{IoU}^l, \\
\hat{R}_{state}^l = \lfloor \frac{R_{state}^l}{v} \rfloor + \frac{Y}{2},
\end{align}
where $v$ denotes the quantization size, and $Y$ denotes the length of the relationship encoding table. We plus $\frac{Y}{2}$ to guarantee that the discrete relative relationship is non-negative.

Next, the discrete relative competitive state $\hat{R}_{state}^l$ is employed as an index to identify the corresponding relationship encoding table $\mathcal{T}^l\in \mathbb{R}^{Y \times d}$ for relative relationship encoding $w^{rel} \in \mathbb{R}^{N' \times N' \times d}$, and a dot product operation is performed with the query vector $v^q \in \mathbb{R}^{N' \times d}$ and key vector $v^k \in \mathbb{R}^{N' \times d}$ in the self-attention as:
\begin{align}
w^{rel}& = \mathcal{T}[\hat{R}_{state}^l], \\ 
rel\_bias_{i,j}& = w^{rel}_{i,j} \cdot v^q_{i} + w^{rel}_{i,j} \cdot v^k_{j},
\end{align}
where $rel\_bias\in \mathbb{R}^{N' \times N'}$ is the relationship bias. It is then added to the self-attention weights, as shown in Figure \ref{figure3}(c).

It is noteworthy that all parameters, with the exception of the requisite relationship encoding table $\mathcal{T}^l$, are shared with \textbf{QCL}. Consequently, the additional computational burden is minimal. In contrast to the direct application of the $MLP$ to obtain the attention weight through the spatial information between queries \cite{EASE-DETR}, our approach represents a novel method of quantifying the spatial and competitive relationships between queries into relative relationship encoding. This facilitates the acquisition of relationship bias through the interaction with semantic features in self-attention, thereby enhancing the robustness of self-attention.

\subsection{Rank Cross Attention (RCA)} 
\label{sec:RCA}
The traditional transformer-based approaches typically compute the similarities of queries $\mathcal{Q}^l \in \mathbb{R}^{N' \times D}$ between the pooling features $F \in \mathbb{R}^{M \times C}$ (ascended from $C$ to $D$ with $MLP$) in the cross-attention as:
\vspace{-1.6mm}
\begin{equation}
Z = \underset{M}{softmax} (\mathcal{Q} \times F^T),
\vspace{-4mm}
\end{equation}
where $Z \in \mathbb{R}^{N' \times M}$ refers to the dot product similarity. The subscript $M$ represents the axis for softmax on the spatial dimension.

We argue that transformer-based instance segmentation differs from standard NLP tasks, with vision tasks cross-attention queries and keys sourced separately. In the traditional cross-attention approach, a query is treated as a set of features, with all key features competing for the query, to calculate the query-key similarity, which determines how the query should incorporate value features. Departing from this convention, we introduce the notion of query competition and propose a new variant: Rank Cross Attention. In Rank Cross Attention, it is postulated that there is competition between queries. For a pooling feature, it should be absorbed by the query that has the highest dot product similarity with it, while other queries should reduce their similarity to this feature, thus increasing the discrepancy between the leading and lagging queries. Consequently, the dot product similarity of queries and pooling features in $N$ dimensions is normalized to ensure that the highest similarity remains unaffected, while other similarities are reduced relatively, according to:
\begin{align}
X &= \mathcal{Q} \times F^T,\\
X_{norm} &= \frac{X - \underset{N}{min}X}{\underset{N}{max}X - \underset{N}{min}X},\\
\hat{Z} &= \underset{M}{softmax} (X \cdot X_{norm}).
\end{align}
\vspace{-5mm}

This design fosters dominance of primary queries in the Rank Cross Attention, empowering them to assimilate richer and denser pooling features, thereby aligning more closely with the ground truth segmentation masks.

\section{Experiments and Analysis}
\begin{table*}[t]
	\centering
	\tabcolsep=0.09cm
	{
		\begin{footnotesize}
			\begin{tabular}{ c | c c | c c c c c c c c c c c c c c c c c c}
				\toprule
				
				Method & mAP & mAP\textsubscript{50} & \rotatebox[origin=c]{90}{bath} & \rotatebox[origin=c]{90}{bed} & \rotatebox[origin=c]{90}{bk.shf} & \rotatebox[origin=c]{90}{cabinet} & \rotatebox[origin=c]{90}{chair} & \rotatebox[origin=c]{90}{counter} & \rotatebox[origin=c]{90}{curtain} & \rotatebox[origin=c]{90}{desk} & \rotatebox[origin=c]{90}{door} & \rotatebox[origin=c]{90}{other} & \rotatebox[origin=c]{90}{picture} & \rotatebox[origin=c]{90}{fridge} & \rotatebox[origin=c]{90}{s. cur.} & \rotatebox[origin=c]{90}{sink} & \rotatebox[origin=c]{90}{sofa} & \rotatebox[origin=c]{90}{table} & \rotatebox[origin=c]{90}{toilet} & \rotatebox[origin=c]{90}{wind.} \\ 
				
				\specialrule{0em}{0pt}{1pt}
				\hline
				\specialrule{0em}{0pt}{1pt}
				
				SPFormer~\cite{spformer} & 54.9 & 77.0 & 74.5 & 64.0 & 48.4 & 39.5 & 73.9 & \underline{31.1} & 56.6 & 33.5 & 46.8 & 49.2 & 55.5 & 47.8 & 74.7 & 43.6 & 71.2 & 54.0 & 89.3 & 34.3 \\
				
				Mask3D~\cite{mask3d} & 56.6 & 78.0 &  \textbf{92.6} & 59.7 & 40.8 & 42.0 & 73.7 & 23.9 & \textbf{59.8} & \underline{38.6} & 45.8 & \textbf{54.9} & \underline{56.8} & \textbf{71.6} & 60.1 & \underline{48.0} & 64.6 & 57.5 & 92.2 & 36.4 \\
				
				MAFT~\cite{maft} & 57.8 &77.4 &77.8 &64.9 &52.0 &\underline{44.9} &76.1 &25.3 &\underline{58.4} &\textbf{39.1} &\textbf{53.0} &47.2 &\textbf{61.7} &49.9 &\textbf{79.5} &47.3 &\underline{74.5} &54.8 &\textbf{96.0} &37.4 \\
				QueryFormer~\cite{queryformer} & \textbf{58.3} & 78.7&  \textbf{92.6} &\underline{70.2} &39.3 &\textbf{50.4} &73.3 &27.6 &52.7 &37.3 &47.9 &\underline{53.4} &53.3 &\underline{69.7} &72.0 &43.6 &\underline{74.5} &\textbf{59.2} &\underline{95.8} &36.3 \\
				
				OneFormer3D~\cite{oneformer3d} &56.6 & \textbf{80.1} &\underline{78.1}&	69.7 &	\underline{56.2} &	43.1 &	\underline{77.0} &	\textbf{33.1} &	40.0 &	37.3 &	\underline{52.9} &	50.4 &	\underline{56.8} &	47.5 &73.2 &	47.0 &	\textbf{76.2} &	55.0 &	87.1 &	\underline{37.9} \\
				\specialrule{0em}{0pt}{1pt}
				\hline
				\specialrule{0em}{0pt}{1pt}
				
				C-SPFormer & \underline{58.0} & \underline{80.0} & 72.1 &  \textbf{70.5} &  \textbf{59.3} & 44.4 &  \textbf{78.6} & 28.6 & 56.4 & 37.6 & 49.8 & \underline{53.4} & 54.6 & 39.0 & \underline{78.5} &  \textbf{57.7} & 70.8 & \underline{57.9} & 95.4 &  \textbf{38.8} \\
				Compare with SPFormer & +3.1 & +3.0 & -2.4 & +6.5 &+10.9&+4.9&+4.7&-2.5&-0.2&+4.1&+3.0&+4.2&-0.9&-8.8&+3.8&+14.1&-0.4&+3.9&+6.1&+4.5\\
				\bottomrule                                   
			\end{tabular}
		\end{footnotesize}
	}
	\vspace{-8pt}
	\caption{3D instance segmentation results on ScanNet \textit{hidden test} set. The term `C-' represents the integration of CompetitorFormer. Methods published before the submission deadline (5 Nov, 2024) are listed. The best results are in bold, and the second best ones are \underline{underlined}.}
	\label{tab:exp_scannet_test}   
\end{table*}
\subsection{Experiments Settings}

\paragraph{Datasets.} 
The experiments are performed on ScanNetv2 \cite{scannet}, ScanNet200 \cite{Rozenberszki_Litany_Dai_2022}, S3DIS \cite{s3dis} and STPLS3D \cite{stpls3d} datasets. The ScanNetv2 dataset \cite{scannet} has 1613 indoor scenes, of which 1201 are used for training, 312 for validation and 100 for testing. The dataset comprises 20 semantic categories and 18 object instances. We report results on both validation and hidden test splits.
ScanNet200 \cite{Rozenberszki_Litany_Dai_2022} extends the original ScanNet semantic annotation with fine-grained categories with the long-tail distribution, resulting in 198 instance with 2 more semantic classes. The training, validation, and testing splits are similar to the original ScanNetv2 dataset.
The S3DIS dataset \cite{s3dis} comprises 6 large-scale regions, encompassing a total of 272 scenes. The instance segmentation task comprises 13 categories. We follow common splits to train on Area[1,2,3,4,6] and evaluate on Area 5. The STPLS3D dataset \cite{stpls3d} is synthetic and outdoor, and closely resembles the data generation process of an aerial photogrammetry point cloud. A total of 25 urban scenes, encompassing 6 $km^2$, have been meticulously annotated with 14 distinct instance classes. The common split \cite{mask3d} is followed to train.  

\paragraph{Baseline.} This study employs state-of-the-art open-source transformer-based frameworks, including SPFormer \cite{spformer}, Mask3D \cite{mask3d}, MAFT \cite{maft} and OneFormer3D \cite{oneformer3d} as experimental baselines. 
Due to the distinctive query selection mechanism inherent to OneFormer3D, only RRE and RCA are incorporated into OneFormer3D.
The SPFormer model is employed for the primary ablation experiments on the ScanNetv2 validation dataset, in order to illustrate the effectiveness of our various components. 
\paragraph{Metrics.} Task-mean average precision (mAP) is utilized as the common evaluation metric for instance segmentation, which averages the scores with IoU thresholds set from $50\%$ to $95\%$, with a step size of $5\%$. Specifically, mAP\textsubscript{50} and mAP\textsubscript{25} denote the scores with IoU thresholds of $50\%$ and $25\%$, respectively. We report mAP, mAP\textsubscript{50} and mAP\textsubscript{25} on all datasets.

\paragraph{Implementation Details.}
CompetitorFormer is integrated into the SPFormer \cite{spformer}, Mask3D \cite{mask3d}, MAFT \cite{maft}, and OneFormer3D \cite{oneformer3d} frameworks, with all training parameters inherited from the original frameworks. The method utilized for query initialisation and the selection of backbones are consistent with that employed in the original framework.
For the SPFormer and MAFT, the number of queries employed is $400$ on ScanNetv2 \cite{scannet} and S3DIS \cite{s3dis}, and $800$ on ScanNet200 \cite{Rozenberszki_Litany_Dai_2022}. For the OneFormer3D, the number of queries is related to the number of superpoints. In the case of the  Mask3D, the number of queries employed is $150$ on ScanNetv2 and S3DIS datasets and $160$ on STPLS3D \cite{stpls3d}.

On ScanNetv2 and ScanNet200, we use a voxel size of 2cm. On S3DIS, voxel size is set to 5cm. As for the STPLS3D, the voxel size is set to 0.33$m$ due to larger scenes. Furthermore, we apply graph-based superpoint clusterization and superpoint pooling on ScanNetv2 and ScanNet200. Following \cite{box2mask}, we get superpoint of S3DIS. Voxels are applied as superpoint on STPLS3D.

\begin{table}[t]
	\centering
	\tabcolsep=0.1cm
	\begin{footnotesize}
		\begin{tabular*}{\linewidth}{@{}c| cccc @{}}
			\toprule
			Method & Venue & mAP & mAP\textsubscript{50} & mAP\textsubscript{25}\\ \midrule
			
			Mask3D$\dagger$~\cite{mask3d} & ICRA23 &  55.2   &   73.7    &  83.5      \\ 
			QueryFormer~\cite{queryformer}& ICCV23 &  56.5   &   74.2    &  83.3  \\
			SPFormer~\cite{spformer} & AAAI23&  56.3   &   73.9    &  82.9    \\
			MAFT~\cite{maft} & ICCV23 &  58.3   &   75.9    &  84.5    \\
			OneFormer3D$\dagger\lhd$~\cite{oneformer3d} & CVPR24 & 58.7 & 77.6 & 85.9\\
			\hline
			C-Mask3D$\dagger$  &   &  56.0 (+0.8)& 73.8 (+0.1)& 83.9 (+0.4)\\
			C-SPFormer  &    &  58.9 (+2.6)& 76.9 (+3.0)&  84.5 (+1.6)  \\
			C-SPFormer$\dagger$  &     &  59.0 (+2.7)& 77.1 (+3.2)& 85.1 (+2.2)  \\
			C-MAFT  &   &  \underline{59.5} (+1.2)& 77.1 (+1.2)& 84.6 (+0.1) \\
			C-MAFT$\dagger$  &    &  \textbf{59.8} (+1.5) & \underline{77.4} (+1.5)& \underline{85.4} (+0.9) \\ 
			C-OneFormer3D$\dagger$ &    &  59.4 (+0.7)& \textbf{77.7} (+0.1)& \textbf{86.2} (+0.3)     \\ 
			\bottomrule   
		\end{tabular*}
	\end{footnotesize}
	\vspace{-8pt}
	\caption{Results of 3D instance segmentation on ScanNetv2 validation set, including comparisons with all current transformer-based methods. The term `C-' represents the integration of CompetitorFormer. $\lhd$ denotes the reproduced result. $\dagger$ denotes that the method uses post-processing. The best results are in bold, and the second best ones are \underline{underlined}.}
	\label{tab:scannet val result}
\end{table}

\begin{table}[t]
	\tabcolsep=0.27cm
	\begin{footnotesize}
		\begin{tabular*}{\linewidth}{c |ccc }
			\toprule
			Method  & mAP & mAP\textsubscript{50} & mAP\textsubscript{25} \\ \midrule
			Mask3D$\dagger$~\cite{mask3d}          &  57.8   &   71.9    & 77.2   \\ 
			QueryFormer~\cite{queryformer}       &  57,7   &   69.9    &  77.1     \\
			SPFormer \cite{spformer}     &  -   &   66.8    &  -    \\
			MAFT$\lhd$ \cite{maft}    &  47.1  &   62.8    &  71.2    \\
			OneFormer3D$\dagger$~\cite{oneformer3d}    &58.7 & 72.0 & -  \\ 
			\hline
			C-Mask3D$\dagger$    &  \underline{58.0} (+0.3) & \underline{72.0} (+0.1) &77.2 (+0.0)  \\
			C-SPFormer     &  52.5 & 68.9 (+2.1)& 77.8  \\			
			C-SPFormer$\dagger$      &  52.6 & 69.2 (+2.4)& \underline{78.1}   \\
			C-MAFT      &  48.1 (+1.0)& 66.3 (+3.5)& 74.1 (+2.9)\\
			C-MAFT$\dagger$  & 48.4  (+1.3)& 67.1  (+4.3)& 74.8 (+3.6)\\ 
			C-OneFormer3D$\dagger$   &  \textbf{59.5} (+0.8) & \textbf{72.9} (+0.9)& \textbf{79.4}     \\ 
			\bottomrule   
		\end{tabular*}
	\end{footnotesize}
	\vspace{-8pt}
	\caption{Results of 3D instance segmentation on S3DIS Area 5, including comparisons with all current transformer-based methods. The term `C-' represents the integration of CompetitorFormer. $\lhd$ denotes the reproduced result. $\dagger$ denotes that the method uses post-processing. The best results are in bold, and the second best ones are \underline{underlined}.}
	\label{tab:S3DIS val result}
\end{table}

\subsection{Main Results}
\paragraph{ScanNetv2}
We present the results of instance segmentation on both the ScanNetv2 test and val sets in Tables \ref{tab:exp_scannet_test} and \ref{tab:scannet val result}, respectively. In hidden test, Competitor-SPFormer increases by $+3.1$ mAP, $+3.0$ mAP\textsubscript{50} compared to baseline, SPFormer \cite{spformer}. Besides, Competitor-SPFormer scores top-2 in the ScanNet hidden test leaderboard at the time of submission with $58.0$ mAP and  $80.0$ mAP\textsubscript{50}, compared with other transformer-based methods. We present segmentation results for 18 classes per method in mAP. Competitor-SPFormer achieves SOTA for five categories, the first in number. In validation set, CompetitorFormer, when combined with four baselines, exhibits improvements of $+0.8$, $+2.7$, $+1.5$ and $+0.7$ mAP, respectively. Furthermore, we present the extra hidden test results of all models just on the training set without post-processing, which can be found in \textbf{Appendix}.
\vspace{-10pt}
\paragraph{S3DIS} 
Table \ref{tab:S3DIS val result} summarizes the results on Area 5 of the S3DIS dataset. Our Method is integrated into OneFormer3D \cite{oneformer3d}, resulting in improving SOTA performance in mAP and mAP\textsubscript{50} for $+0.8$ and $+0.9$, respectively. Furthermore, it enhances the remaining three frameworks to varying degrees. The number of initial queries of Mask3D is less than that of the other three frameworks, and the degree of competition between queries is naturally limited, which results in less significant improvements.
\vspace{-10pt}
\paragraph{ScanNet200}
Table \ref{tab:scores_scannet200} illustrates the quantitative result on ScanNet200. Our proposed method outperforms the second best performing method with margins of $+1.5$ and $+0.9$ in mAP, mAP\textsubscript{50}, improving the performance of SOTA $4.9\%$ and $2.2\%$ respectively. 
\vspace{-10pt}
\paragraph{STPLS3D}  Table \ref{tab:scores_stpls3d} shows the quantitative comparison on the validation set of STPLS3D dataset. Our method outperforms all existing methods, improving SOTA performance in mAP and AP50 for $+0.2$ and $+0.1$, respectively. The limited number of queries in Mask3D and the sparse query distribution limit the performance of ComeptitorFormer.

\begin{figure*}[t]
	\centering
	\begin{subfigure}[b]{0.33\linewidth}
		\centering
		\includegraphics[width=5cm,height=3.75cm]{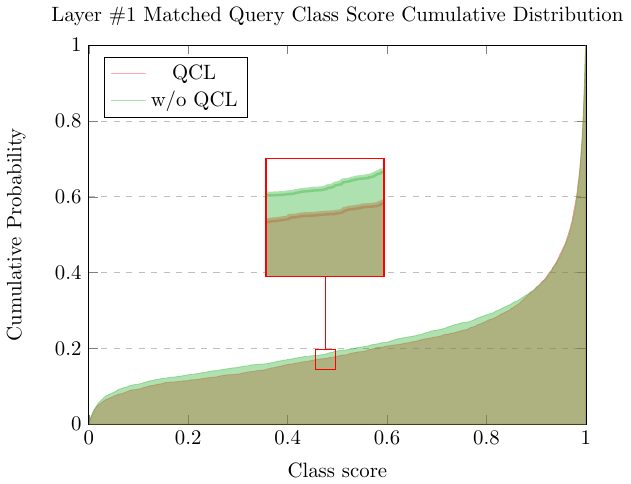}
		\caption{}
		\label{fig:4a}
	\end{subfigure}
	\begin{subfigure}[b]{0.33\linewidth}
		\centering
		\includegraphics[width=5cm,height=3.75cm]{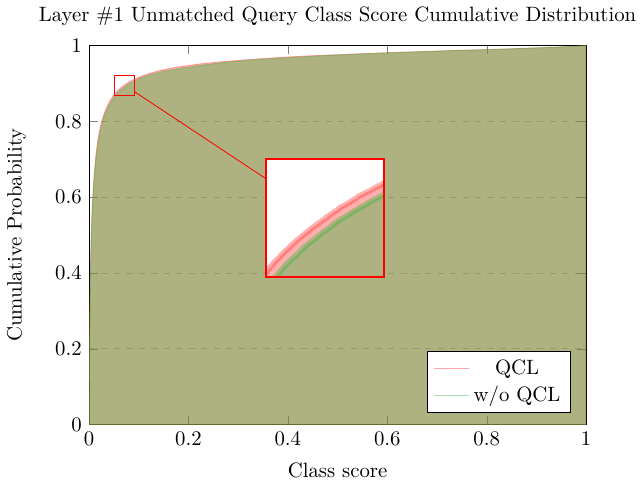}
		\caption{}
		\label{fig:4b}
	\end{subfigure}
	\begin{subfigure}[b]{0.33\linewidth}
		\centering
		\includegraphics[width=5cm,height=3.75cm]{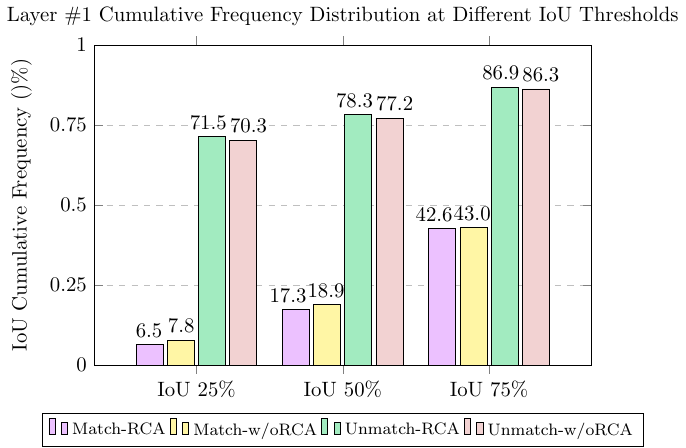}
		\caption{}
		\label{fig:4c}
	\end{subfigure}
	\vspace{-20pt}
	\caption{(a)\&(b) Cumulative distribution of the class scores on the matched (unmatched) queries with or without QCL, (c) Cumulative distribution of the IoU scores on the matched (unmatched) queries with or without RCA.}
	\label{figure4}
\end{figure*}

\begin{table}[t]
	\centering
	
	\begin{minipage}{.48\linewidth}
		\centering
		\begin{footnotesize}
			\tabcolsep=0.05cm
			\begin{tabularx}{\linewidth}{ccc}
				\toprule
				& \multicolumn{2}{c}{\textbf{ScanNet 200}}\\
				\cmidrule{2-3}
				Method                        & mAP & mAP$_{50}$ \\ \midrule
				SPFormer \cite{mask3d} & 25.2 & 33.8\\
				MAFT\cite{maft} & 29.2 & 38.2 \\
				OneFormer3D$\dagger$~\cite{oneformer3d}    &\underline{30.6} & \underline{40.8}   \\ 
				\hline
				C-SPFormer$\dagger$ & 30.2 & 38.8 \\
				C-MAFT$\dagger$ & \textbf{32.1} & \textbf{41.7} \\
				\bottomrule
			\end{tabularx}
		\end{footnotesize}
		\subfloat[]{\label{tab:scores_scannet200}}
	\end{minipage}%
	\hfill
	\begin{minipage}{.48\linewidth}
		\centering
		\begin{footnotesize}
			\tabcolsep=0.00cm
			\begin{tabularx}{\linewidth}{ccc}
				\toprule
				& \multicolumn{2}{c}{\bf STPLS3D}\\
				\cmidrule{2-3}
				Method                          &     mAP  & mAP$_{50}$ \\ \midrule
				PointGroup \cite{pointgroup}  &         23.3 & 38.5\\
				ISBNet \cite{ngo2023isbnet} & 49.2 & 64.0 \\
				Mask3D $\dagger$ \cite{mask3d}         &   \underline{57.3} & \underline{74.3}\\
				Spherical Mask$\dagger$ ~\cite{sphericalmask}     &         52.2 &  68.3\\
				\hline
				C-Mask3D$\dagger$              & \textbf{57.5}    & \textbf{74.4} \\
				\bottomrule
			\end{tabularx}
		\end{footnotesize}
		\subfloat[]{\label{tab:scores_stpls3d}}
	\end{minipage}
	\vspace{-18pt}
	\caption{
		3D Instance Segmentation Scores on ScanNet200 \cite{Rozenberszki_Litany_Dai_2022} and STPLS3D \cite{stpls3d}. We report mean average precision (mAP) with different IoU thresholds over 14 classes on the STPLS3D validation set. The term `C-' represents the integration of CompetitorFormer. $\dagger$ denotes that the method uses post-processing. The best results are bold, and the second best ones are in \underline{underlined}.
	}
\end{table}

\subsection{Ablation Studies}
The systematic analysis was performed to assess the impact of each proposed component in our approach.
A step-by-step approach was employed, whereby modules were added incrementally to the baseline (Table \ref{tab:F5A}), then merged into the baseline (Table \ref{tab:F5B}), and finally removed from the approach (Table \ref{tab:F5C}).
This process allowed us to understand the impact of each individual component on the final performance. In addition, we performed statistical and quantitative analyses to fully assess the functionality of each component.

\paragraph{Query Competitor Layer (QCL).} 
The QCL mechanism effectively integrates the competition of queries into SPFormer, compensating for the lack of handling the competitive relationship of queries in the self-attention. The performance of segmentation is continuously improved by utilizing QCL ($+1.3$ mAP when adding QCL to the SPFormer; $+1.1$ mAP when completing our approach). Furthermore, the classification scores of matched and unmatched queries in bipartite matching are counted after each decoder layer, and the cumulative probability distribution is calculated. Figure \ref{fig:4a} illustrates that the matched query scores generated by our method exhibit a marked enhancement in comparison to SPFormer, with a notable reduction in accumulation at low scores (green area). In contrast, Figure \ref{fig:4b} illustrates that the unmatched query scores produced by our method are more concentrated at low scores (red area). It has been demonstrated that QCL can effectively mitigate competition between queries by enhancing the classification accuracy of matched queries and suppressing the classification scores of unmatched queries. A more comprehensive account can be found in \textbf{Appendix}.

\begin{table}[t]
	\centering
	\begin{footnotesize}
		\begin{subtable}{\columnwidth}
			\centering
			
			\begin{subtable}{\columnwidth}
				\centering	
				\begin{tabular}{@{}ccc|ccc@{}}
					\toprule
					QCL & RRE & RCA & mAP & mAP\textsubscript{50} & mAP\textsubscript{25} \\ \midrule
					\ding{55} & \ding{55} & \ding{55} &  56.3   &   73.9    &  82.9      \\
					\ding{51} & \ding{55} & \ding{55} &  57.6 (+1.3)  & 75.0 (+1.1)  &  83.7 (+0.8)     \\
					\ding{51} & \ding{51} & \ding{55} &  58.3 (+2.0)  & 76.2 (+2.3)  &  84.1 (+1.2)     \\
					\ding{51} & \ding{51} & \ding{51} &  \textbf{58.9 (+2.6)} & \textbf{76.9 (+3.0)} & \textbf{84.5  (+1.6)}   \\  \bottomrule
				\end{tabular}
				
				\caption{ Effect of gradually adding modules on the baseline.}
				\label{tab:F5A}
			\end{subtable}

			\begin{tabular}{@{}ccc|ccc@{}}
				\toprule
				QCL & RRE & RCA & mAP & mAP\textsubscript{50} & mAP\textsubscript{25} \\ \midrule
				\ding{55} & \ding{55} & \ding{55} &  56.3   &   73.9    &  82.9       \\
				\ding{51} & \ding{55} & \ding{55} &  \textbf{57.6 (+1.3)}   & 75.0 (+1.1) &  83.7 (+0.8)    \\
				\ding{55} & \ding{51} & \ding{55} &  57.4 (+1.1)  & \textbf{75.3 (+1.4)}    &  83.9 (+1.0) \\
				\ding{55} & \ding{55} & \ding{51} &  57.3 (+1.0)  & 74.9 (+1.0)  &  \textbf{84.0 (+1.1)}   \\  \bottomrule
			\end{tabular}
			\caption{ Effect of incorporating each module on the baseline.}
			\label{tab:F5B}
		\end{subtable}
		

		
		\begin{subtable}{\columnwidth}
			\centering
			\begin{tabular}{@{}ccc|ccc@{}}
				\toprule
				QCL & RRE & RCA & mAP & mAP\textsubscript{50} & mAP\textsubscript{25} \\ \midrule			\ding{51} & \ding{51} & \ding{51} &  \textbf{58.9 (+2.6)} & \textbf{76.9 (+3.0)} & \textbf{84.5  (+1.6)}  \\
				\ding{55} & \ding{51} & \ding{51} &  57.8  & 75.9  &  84.4   \\
				\ding{51} & \ding{55} & \ding{51} &  58.0  & 75.7  &  84.2     \\
				\ding{51} & \ding{51} & \ding{55} &  58.3  & 76.2  &  84.1   \\  \bottomrule
			\end{tabular}
			\caption{ Effect of removing each module in our method.}
			\label{tab:F5C}
		\end{subtable}
	\end{footnotesize}
	\vspace{-12pt}
	\caption{Ablation experiments based on SPFormer. QCL: query competitor layer. RRE: relative relationship encoding. RCA: rank cross attention.}
	\label{tab:F5}
\end{table}
\begin{figure*}[h]
	\centering
	\includegraphics[width=0.25\columnwidth]{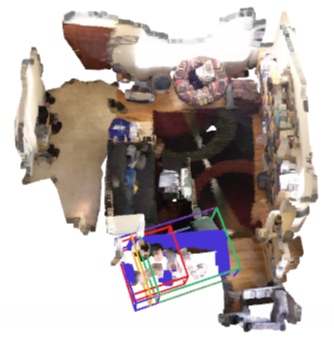} \quad
	\includegraphics[width=0.25\columnwidth]{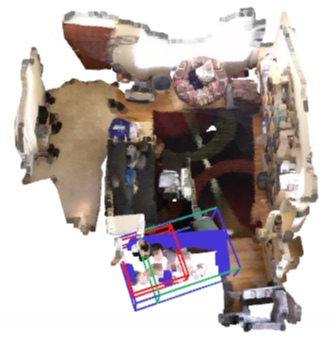} \quad
	\includegraphics[width=0.25\columnwidth]{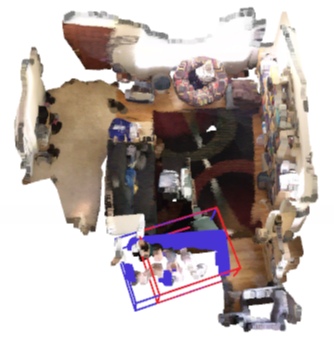} \quad 
	\includegraphics[width=0.25\columnwidth]{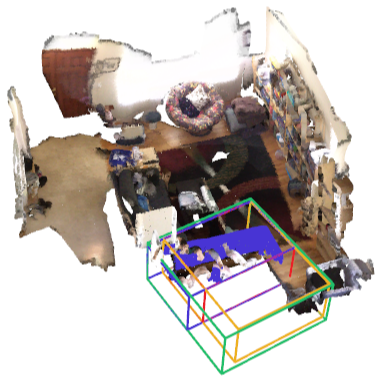} \quad
	\includegraphics[width=0.25\columnwidth]{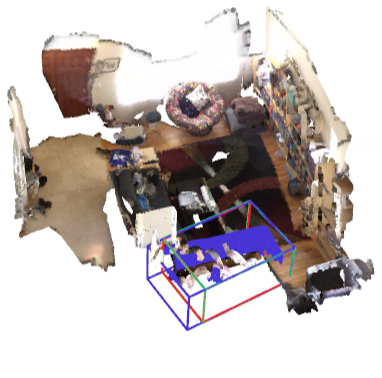} \quad
	\includegraphics[width=0.25\columnwidth]{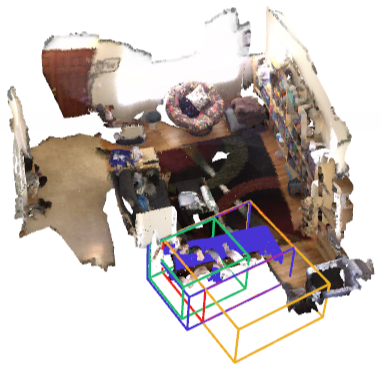} \\
	\includegraphics[width=0.25\columnwidth]{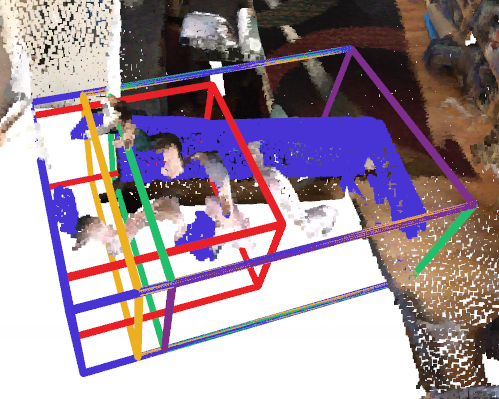} \quad
	\includegraphics[width=0.25\columnwidth]{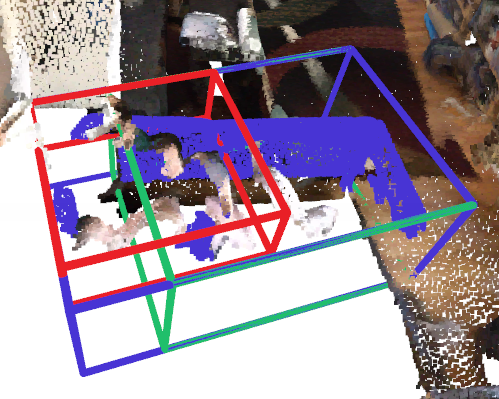} \quad
	\includegraphics[width=0.25\columnwidth]{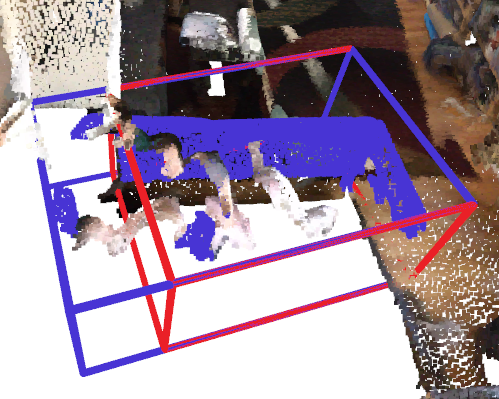} \quad 
	\includegraphics[width=0.25\columnwidth]{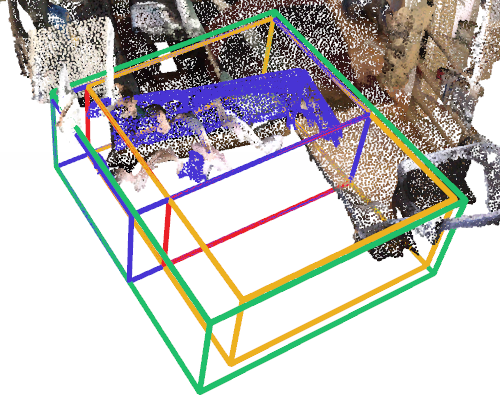} \quad
	\includegraphics[width=0.25\columnwidth]{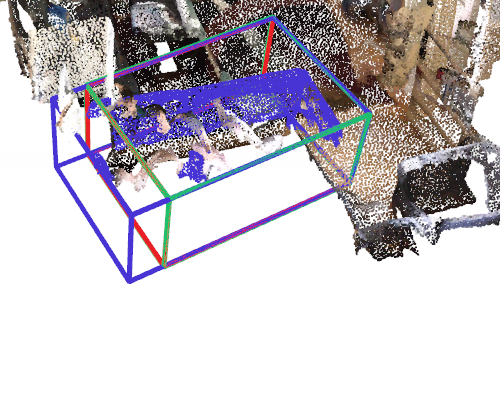} \quad
	\includegraphics[width=0.25\columnwidth]{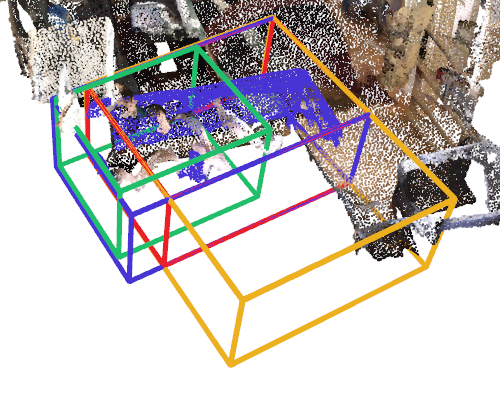} \\
	\vspace{-6pt}
	\caption{Visualisation of competitive query in 1st/3rd/5th decoder layers for both the Competitor-SPFormer (left 3 columns) and SPFormer (right 3 columns). The depiction employs 3D bounding boxes to denote predicted instances. This scenario depicts queries in which the IoU is greater than $0.25$. The color of each fraction matches the corresponding bounding box's color, where the blue box is the ground truth box and the rest of the colored boxes (e.g., red, green, etc.) are the prediction boxes.}
	\label{figure-QA}
\end{figure*}

\paragraph{Relative Relationship Encoding (RRE).}
Table \ref{tab:F5B} and \ref{tab:F5C} shows that RRE can
improve the mAP ($+1.1$ mAP when adding RRE to the SPFormer baseline, comparing row1 and row3 in Table \ref{tab:F5B};
$+0.9$ mAP when adding RRE to complete our approach, comparing row1 and row3 in Table \ref{tab:F5C}). A comparison of the second and third rows of Table \ref{tab:F5A} reveals that RRE continues to demonstrate $+0.7$ mAP when parameters are shared with QCL. Furthermore, RRE improves mAP\textsubscript{50} more than mAP.

\paragraph{Rank Cross Attention (RCA).} 
Table \ref{tab:F5} also demonstrates the efficacy of the proposed RCA module by comparing row
1 and row 4 in Table \ref{tab:F5B}. The incorporation of RCA alone enhances the mAP of the SPFormer baseline by $+1.0\%$, while
the integration of RCA into our comprehensive approach further elevates mAP by $+0.6\%$, as illustrated in row 1 and
row 4 of Table \ref{tab:F5C}. Furthermore, we investigate the impact of the RCA module on the query IoU. The IoU of each unmatched query is defined as the IoU of between it and the corresponding ground truth mask with the minimum cost among the bipartite matching. Figure \ref{fig:4c} illustrates the cumulative distribution of query IoU across three distinct IoU thresholds. With respect to the aforementioned three thresholds, the cumulative distribution of matched query IoU of our method is observed to be lower than SPFormer, whereas the cumulative distribution of unmatched query IoU is higher than SPFormer. The incorporation of RCA enhances the IoU between the aligned query and the target instance mask, thereby bringing the positive query closer to the ground truth. The IoU between the unmatched query and the ground truth mask is effectively pushed away from the ground truth mask by RCA. This demonstrates that the RCA module is an effective means of mitigating the competition phenomenon between queries. A more comprehensive account can be found in \textbf{Appendix}.

\subsection{Qualitative Analysis}
To provide a deeper understanding of the inter-query competition, Figure \ref{figure-QA} presents visual comparisons of instance segmentation outputs along on SPFormer and Competitor-SPFormer between the 1st/3rd/5th decoder layers. It is noteworthy that our method, as illustrated on the left, effectively alleviates query competition when compared to the baseline on the right. While both methods exhibit multiple queries contending for similar instances in the decoder's initial stages, our approach promptly suppresses less competitive queries while allowing more competitive ones to advance iteratively through the decoder layer and  continuous optimization. These qualitative observations highlight how our strategy improves performance by effectively reducing inter-query competition.

\section{Conclusion}
In this work, we have introduced the CompetitorFormer, a series of simple yet effective plug-and-play competition-oriented designs to address the task of 3D transformer-based instance segmentation. 
The core insights behind its effectiveness are (1) to mitigate competition by enhancing classification confidence for matched positive queries and attenuating classification confidence for unmatched queries, and (2) to enhance the ability of the dominant query to absorb object features and suppress the rest of the queries to increase the discrepancy among queries.
Extensive experiments on ScanNetv2, S3DIS, ScanNet200 and STPLS3D datasets show that our method achieves robust and significant performance gain on all datasets, surpassing state-of-the-art approaches in 3D instance segmentation.

Our method is not without limitations. For example, our QCL module is only compatible with frameworks where the number of queries remains constant and cannot be more effectively integrated with the existing SOTA method, OneFormer3D. Consequently, the potential for enhancement in OneFormer3D is constrained. Addressing this limitation might lead to improvement in future work. Additionally, extending our method to more fields such as 3D object detection and 3D panoptic segmentation would be an interesting research topic.

{
    \small
    \bibliographystyle{ieeenat_fullname}
    \bibliography{main}
}


\end{document}